\documentclass[10pt,twocolumn,letterpaper]{article}
\pdfoutput=1
\usepackage[pagenumbers]{cvpr} 
\usepackage{algorithm}
\usepackage{algorithmic}
\usepackage{indentfirst}
\usepackage[dvipsnames]{color}
\usepackage[dvipsnames]{xcolor}

\definecolor{cvprblue}{rgb}{0.21,0.49,0.74}
\usepackage[pagebackref,breaklinks,colorlinks,citecolor=cvprblue]{hyperref}

\def\paperID{9026} 
\def\confName{CVPR}
\def\confYear{2024}

\title{Tuning-Free Noise Rectification for High Fidelity Image-to-Video Generation}

\author{Weijie Li, ~~%
    Litong Gong, ~~%
    Yiran Zhu, ~~%
    Fanda Fan, ~~%
    Biao Wang, ~~%
    Tiezheng Ge, ~~%
    Bo Zheng ~~%
    \\
Alimama Tech, Alibaba Group\\
Beijing, China\\
{\tt\small \{weijie.lwj0, gonglitong.glt, yizhu.zyr, fanda.ffd, }\\
{\tt\small eric.wb, tiezheng.gtz, bozheng\}@alibaba-inc.com}}

\begin{document}
\maketitle
\begin{abstract}
Image-to-video (I2V) generation tasks always suffer from keeping high fidelity in the open domains. 
Traditional image animation techniques primarily focus on specific domains such as faces or human poses, making them difficult to generalize to open domains. 
 Several recent I2V frameworks based on diffusion models can generate dynamic content for open domain images but fail to maintain fidelity. 
We found that two main factors of low fidelity are the loss of image details and the noise prediction biases during the denoising process. 
To this end, we propose an effective method that can be applied to mainstream video diffusion models. 
This method achieves high fidelity based on supplementing more precise image information and noise rectification. 
Specifically, given a specified image, our method first adds noise to the input image latent to keep more details, then denoises the noisy latent with proper rectification to alleviate the noise prediction biases. 
Our method is tuning-free and plug-and-play. 
The experimental results demonstrate the effectiveness of our approach in improving the fidelity of generated videos. For more image-to-video generated results, please refer to the project website: \href{https://noise-rectification.github.io/}{https://noise-rectification.github.io/}.

\end{abstract}
    
\section{Introduction}
With the remarkable breakthroughs of diffusion models in generating exquisite images~\cite{5_DiffusionBeatGAN_DhariwalN21, 7_SD_RombachBLEO22, 10_DALLE2_ramesh2022hierarchical, 11_Imagen_SahariaCSLWDGLA22,47_DreamBooth_ruiz2023dreambooth}, researchers are exploring the further potential of diffusion models to achieve more coherent video generation. 
Some recent works~\cite{18_VDM_HoSGC0F22, 19_Make-A-Video_SingerPH00ZHYAG23, 20_ImagenVideo_ho2022imagen, 23_ModelScope_wang2023modelscope, 36_VideoComposer_2023videocomposer} have made incremental progress in the text-to-video (T2V) task to generate videos that align with the input text. However, a textual description can correspond to various imaginable videos, which may not necessarily meet people’s specific expectations. Therefore, the reference image is proposed to guide the video generation process, aiming to generate videos that closely align with the given image or even strictly start from the still image, which is called the image-to-video (I2V) task. 

The concept of image-to-video is not novel and has long existed in traditional computer vision tasks, such as facial animation~\cite{48_face_vid2vid_wang2021one,49_FOMM_siarohin2019first}, body motion synthesis~\cite{50_LFDM_ni2023conditional}, nature-driven animation~\cite{51_fluid_mahapatra2022controllable, 52_Eulerian_holynski2021animating}, and video prediction~\cite{53_Flow_video_prediction_li2018flow, 54_latent_video_prediction_franceschi2020stochastic, 55_diffusion_video_prediction_hoppe2022diffusion}, which can all be considered as I2V tasks. However, these tasks were either limited to specific domains (such as faces, human poses, and simple natural scenes) or focused on relatively simple scenarios (such as animating fonts~\cite{54_latent_video_prediction_franceschi2020stochastic}, drawing~\cite{56_animateDrawing_smith2023method} or moving rigid objects~\cite{57_MakeItMove_hu2022make}). The proposed solutions for these specific tasks are difficult to be applied to open-domain images. Moreover, previous studies~\cite{51_fluid_mahapatra2022controllable, 18.0_RVD_yang2022diffusion, 57_MakeItMove_hu2022make, 55_diffusion_video_prediction_hoppe2022diffusion, 58_MCVD_voleti2022mcvd, 59_CCVS_le2021ccvs} adopted the autoregressive approach to generate the video sequences, which is computationally expensive and still faces challenges in complex open-domain scenarios. Recently, the emerged diffusion models have demonstrated strong generative capabilities and significant extensibility by learning the data distribution from noise. As a video can be considered as a temporal sequence (batch)  of highly correlated images, it is feasible to process videos in batches using diffusion models. Consequently, there is a growing focus on leveraging diffusion models for image-to-video task, attracting significant attention from both research and industry. 

However, current I2V research~\cite{34_Seer_gu2023seer, 35_VideoCrafter_chen2023videocrafter1, 36_VideoComposer_2023videocomposer, 37_I2VGen-XL_zhang2023i2vgenxl}
primarily relies on enhancing the supervision of image signals to guide the video diffusion model. As a result, the generated videos are only able to resemble the given image. In our view, existing video diffusion models in these works already exhibit strong capabilities for generating dynamic motions, but they struggle to maintain fidelity, which can be attributed to two main factors. One is the loss of image details, such as adopting IP-Adapter~\cite{14_IP-Adapter_ye2023ip-adapter} or ControlNet~\cite{12_ControlNet_zhang2023adding} only extracts partial image representation. Another is the noise prediction biases during the denoise process, due to the unattainable perfect zero loss in training the video diffusion model, even when the entire image information has been injected or concatenated. Inspired by the transition refinement with
the pivotal noise vector in recent image editing work~\cite{15_ILVR_ChoiKJGY21, 16_SDEdit_MengHSSWZE22, 17_Null-text_MokadyHAPC23}, we propose to set the direction of initial noise as the pivotal reference in the denoising process. Specifically, we design an image-to-video pipeline, which adopts the ``noising and rectified denoising" reverse process to improve the fidelity of generated video. Our method utilizes pre-trained video latent diffusion models (VLDM) to generate fluent motion between frames. During the inference, we first add initial noise to the given image to alleviate the loss of detail information, denoted as ``noising" stage. Then we properly rectify the predicted noise using the pivotal reference noise in the reverse denoise timesteps to alleviate the noise prediction biases, denote as ``rectified-denoising" stage. 
Additionally, in order to control the  retention degree of the reference image, we further introduce a practical step-adaptive intervention strategy based on noise rectification.

In general, we propose an effective method to utilize the existing pre-trained video diffusion models for image-to-video tasks. The comparison experiments with current public I2V works and several I2V attempts in the active community have demonstrated the effectiveness of our methods in generating videos with higher fidelity. Moreover,  our method does not require extra training and is simple to implement, which can be seamlessly integrated with current pre-trained open-domain video diffusion models in a plug-and-play manner, enabling high-fidelity I2V generation in open domains.
\section{Related Work}
The diffusion models have achieved great success in generative tasks in recent years. Due to the high correlation between image and video modalities, many ideas and insights in current video generation work have been inspired by extensive image generation work. Therefore, we introduce the related work on image and video generation here.

\subsection{Image Generation with Diffusion Model}
Compared to the traditional GAN~\cite{1_GAN_goodfellow2014generative} and VAE~\cite{3_VAE_kingma2013auto} based methods, the diffusion models ~\cite{2_diffusion_sohl2015deep, 4_DDPM_HoJA20, 5_DiffusionBeatGAN_DhariwalN21, 45_SDE_song2020score, 46_DDIM_song2020denoising, 10_DALLE2_ramesh2022hierarchical, 11_Imagen_SahariaCSLWDGLA22} have demonstrated more powerful capabilities to produce high-quality images with realistic textures and fine details.  The U-Net~\cite{6_U-Net_RonnebergerFB15} with the attention layer is the widely adopted structure in image diffusion models to predict noise. To save computation costs, Stable Diffusion (SD)~\cite{7_SD_RombachBLEO22} proposed the latent diffusion model (LDM), which utilized VAE~\cite{3_VAE_kingma2013auto} to encode the image into a latent space and perform the diffusion process on the latent space. To enhance the controllability and support various control conditions such as depth, reference image, normal map and canny map, ControlNet~\cite{12_ControlNet_zhang2023adding} and T2I-Adapter~\cite{13_T2I-Adapter_mou2023t2iadapter} introduced a flexible adapter based on the SD~\cite{7_SD_RombachBLEO22} for controlled generation. Recently, IP-Adapter~\cite{14_IP-Adapter_ye2023ip-adapter} also proposed an image prompt adapter for T2I models to guide the image generation with the reference image.  Besides, SDEdit~\cite{16_SDEdit_MengHSSWZE22} added noise to the input stroke image and progressively denoised the resulting image to increase the realism of the synthesized image. In our image-to-video work, we adopt an inflated 3D U-Net similar to the T2I task and our noise rectification also takes inspiration from the transition refinement in the image editing works~\cite{15_ILVR_ChoiKJGY21, 16_SDEdit_MengHSSWZE22, 17_Null-text_MokadyHAPC23}.

\subsection{Video Generation with Diffusion Model}
Thanks to the significant progress of text-to-image generation, video generation has also started to develop from the text-to-video (T2V) task. VDM~\cite{18_VDM_HoSGC0F22} introduced a pioneering video diffusion model that extends the 2D U-Net to a 3D U-Net structure, jointly training both images and videos in the pixel space. Subsequent methods~\cite{21_MagicVideo_zhou2023magicvideo, 22_LVDM_he2022lvdm, 23_ModelScope_wang2023modelscope, 24_Latent-Shift_an2023latentshift, 25_animatediff_guo2023animatediff, 26_Show-1_zhang2023show, 27_Align-Your-Latents_BlattmannRLD0FK23, 28_LAVIE_wang2023lavie, 29_VideoFusion_luo2023videofusion} mostly adopted the latent space to reduce memory requirements and speed up the training and inference. To optimize the running time required for video generation, most works (Make-A-Video~\cite{19_Make-A-Video_SingerPH00ZHYAG23}, ModelScopeT2V~\cite{23_ModelScope_wang2023modelscope}, Latent-Shift~\cite{24_Latent-Shift_an2023latentshift}, AnimateDiff~\cite{25_animatediff_guo2023animatediff}) were built upon the pre-trained T2I models and incorporated temporal modules, enabling batch generation of all video frames simultaneously. Particularly, AnimateDiff~\cite{25_animatediff_guo2023animatediff} only trains a motion module that can be adapted to various personalized T2I models. Text2Video-Zero~\cite{33_Text2Video-Zero_khachatryan2023text2videozero} proposed a training-free sampling method to enrich motion dynamics and maintain temporal consistency with cross-frame attention. Besides, the cascade framework of video diffusion models is also used to generate high-resolution~\cite{20_ImagenVideo_ho2022imagen, 27_Align-Your-Latents_BlattmannRLD0FK23, 28_LAVIE_wang2023lavie} and longer videos~\cite{22_LVDM_he2022lvdm, 32_NUWA-XL_YinWYWWNYLL0FGW23}.

Similar to image generation, introducing more control conditions in video generation is also crucial. Recently, to make the generated videos more controllable, recent work has introduced various conditions into the video diffusion models, including depth~\cite{19_Make-A-Video_SingerPH00ZHYAG23, 38_Gen-1_esser2023structure}, pose~\cite{40_FollowYourPose_ma2023follow, 41_DreamPose_karras2023dreampose}, guided motion from trajectory~\cite{42_DragNUWA_yin2023dragnuwa}, stroke painting~\cite{43_MCDiff_chen2023motionconditioned} or frequency analysis ~\cite{44_GenerativeImageDynamics_li2023generative}. As to the image condition,  existing video generation work mainly draws on experiences from the image generation field, e.g., enhancing the image guidance using ControlNet~\cite{12_ControlNet_zhang2023adding} and IP-Adapter~\cite{14_IP-Adapter_ye2023ip-adapter}. Besides, Seer~\cite{34_Seer_gu2023seer} concatenated the conditional image latent with the noisy latent and employed causal attention in the temporal module of 3D U-Net for image-to-video tasks. VideoComposer~\cite{36_VideoComposer_2023videocomposer} proposed to concatenate the image embedding with the noisy latent along the feature channel, as well as support forwarding the style of the given image into the video latent diffusion model (VLDM).  Recently, VideoCrafter~\cite{35_VideoCrafter_chen2023videocrafter1} extracted the image feature into the VLDM for the image-to-video task. Similarly, I2VGen-XL~\cite{37_I2VGen-XL_zhang2023i2vgenxl} both added the image latent with the noisy latent in the input layer and built a global encoder to extract the image CLIP feature into the VLDM. However,  these image-to-video works either have limited fidelity or require fine-tuning the whole VLDM. In comparison, our noise rectification method is tuning-free and maintains high fidelity.

\section{Preliminary}

\begin{figure}
\centering
\includegraphics[width=\linewidth,scale=1.0]{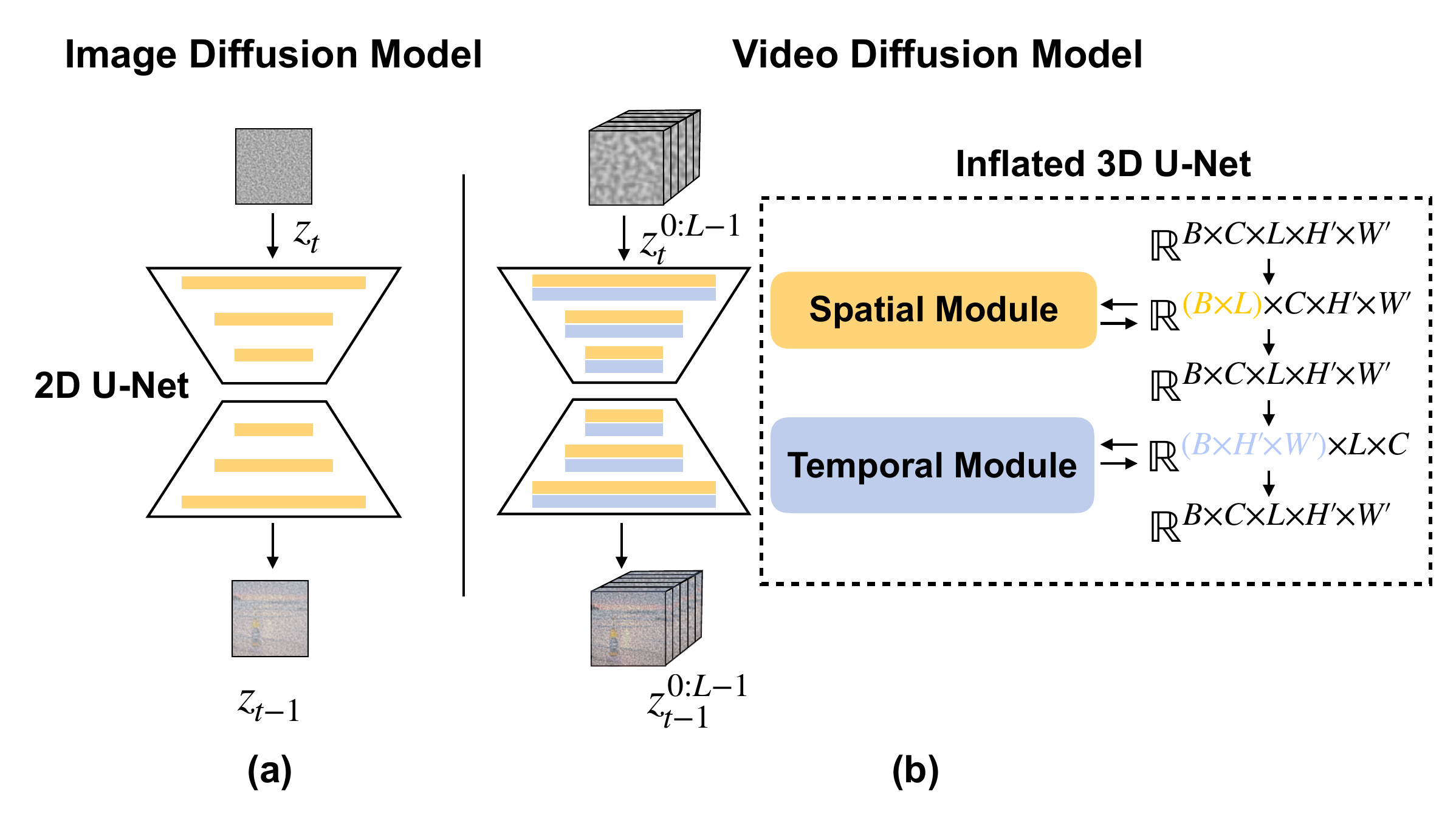}
\caption{The general framework of image diffusion model and video diffusion model with inflated 3D U-Net structure.}
\label{fig:image_vs_video}
\end{figure}

\subsection{Image-to-Video Task Definition}\label{Image-to-Video Task Definition}
All video generation tasks require generating coherent frames that maintain both visual consistency and logical motion. Specifically, image-to-video (I2V) task is defined as generating a video from a specified reference image. Its goal is to transform the static nature of an image into the dynamic visual representation, adding motion and fluidity to the content. Compared to the text-to-video (T2V) task, I2V prioritizes high fidelity with the conditional image, while dynamic motion in the video can be learned through common prior knowledge or driven by the given conditions like the text description or other data forms. Here we focus on the text-conditioned image-to-video task, and this definition can be formulated as, given a still image $I$ and a text description $c$, the generative system outputs a predicted video ${V}^{0:L-1} = \left\{ \bar{I}^0, \dots, \bar{I}^{L-1} \right\}$, where $L$ represents the video length. The objective is to keep appearance consistent with the given initial image $I$, as well as ensure the generated video aligns with the text description $c$.

\subsection{Video Latent Diffusion Models}
The diffusion models~\cite{2_diffusion_sohl2015deep, 4_DDPM_HoJA20,45_SDE_song2020score, 46_DDIM_song2020denoising}  are a class of generative models inspired by non-equilibrium thermodynamics, which define the Markov chain to perturb data to noise in the diffusion process and then learn to convert noise back to data in the reverse process. Formally, in the diffusion process, given a data distribution $z_0 \sim q(z_0)$, the forward Markov chain gradually adds Gaussian noise to the data sample in the $T$ timesteps, thus obtaining a sequence of noisy data $\left\{z_1, z_2, \dots, z_T\right\}$ conditioned on $z_0$, following the transition formula, which can be denoted as:
\begin{align}
    q(z_{1:T}|z_0) &= \prod \limits_{t=1}^T q(z_t|z_{t-1}), \\
    q(z_t|z_{t-1}) &= \mathcal{N}(z_t; \sqrt{\alpha_t}z_{t-1}, (1-\alpha_t)\mathbf{I})\label{eq:2},
\end{align}
where $\left\{\alpha_t \in (0,1)\right\}^T_{t=1}$ is a variance schedule to control the step size. In the reverse process, a model $p_{\theta}$ is learned to denoise from the noisy prior $z_T \sim \mathcal{N}(\mathbf{0}, \mathbf{I})$ to gradually generate the desired data iteratively following:
\begin{equation}
    p_\theta(z_{t-1}|z_t) = \mathcal{N}(z_{t-1}; \mu_{\theta}(z_t, t), \Sigma_\theta(z_t, t)),
\end{equation}
where $\theta$ is the model parameters, $\mu_{\theta}(z_t, t)$ and $\Sigma_{\theta}(z_t, t)$ denote the predicted mean and variance by the model. 

In the image generative tasks, the denoising model is usually designed as the U-Net network architecture and learned with the objective function 
\begin{equation}
    \min \limits_\theta \mathbb{E}_{z_0 \sim p_{data}, t, \epsilon \sim \mathcal{N}(\mathbf{0},\mathbf{I})} [{\Vert \epsilon - \epsilon_\theta(z_t, c, t) \Vert}^2_2],
\end{equation}
where $\epsilon$ and $\epsilon_\theta$ are the actual and predicted noise respectively, $c$ represents various conditions like text, image, or other control signals. Furthermore, to reduce the computational complexity, diffusion models are utilized in a lower-dimensional latent space rather than the pixel space, which is denoted as the latent diffusion model~\cite{7_SD_RombachBLEO22}.

\begin{figure}[!t]
\centering
\includegraphics[width=\linewidth,scale=1.0]{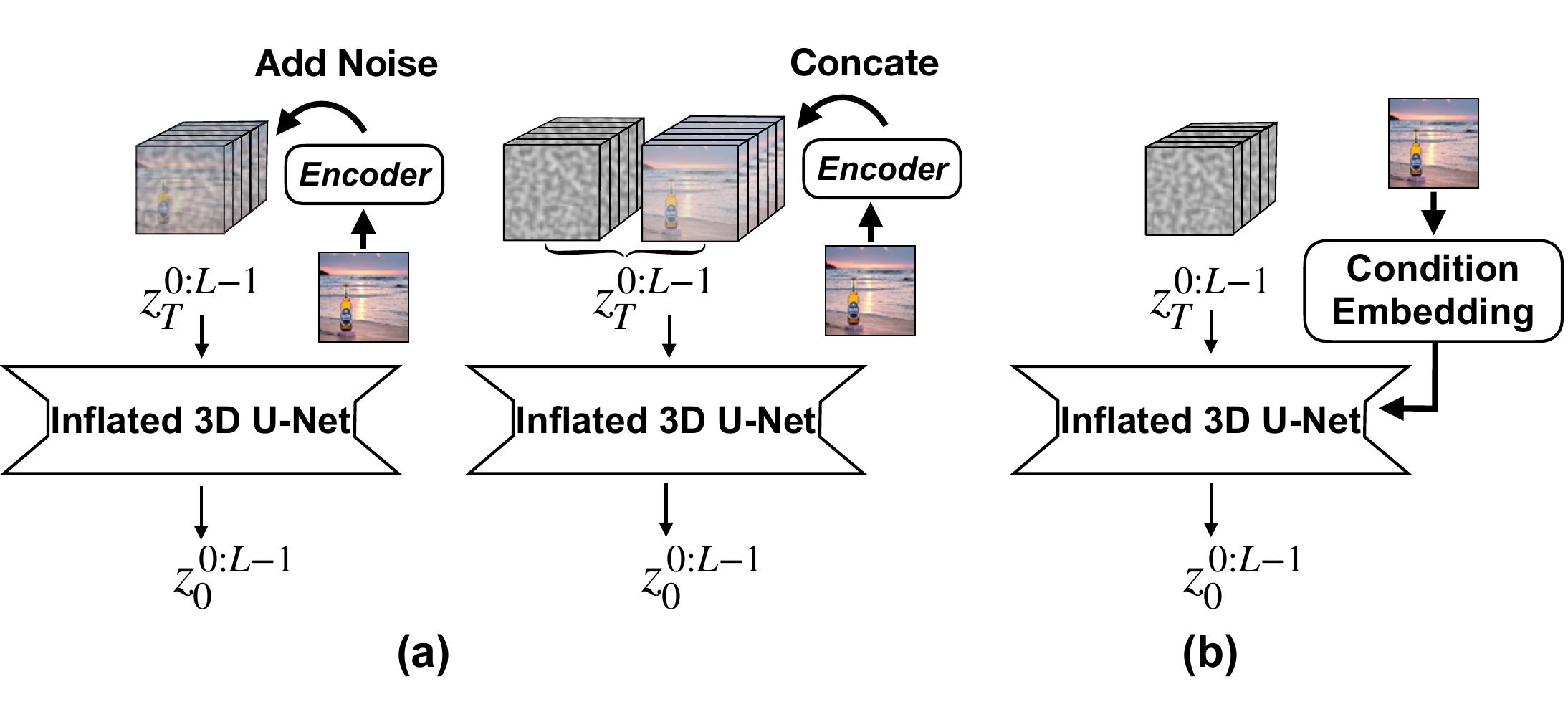}
\caption{Two basic approaches in existing research and community regarding image-to-video generation.}
\label{fig:two_I2V_ways}
\end{figure}

\begin{figure*}[!ht]
\centering
\includegraphics[width=0.75\linewidth]{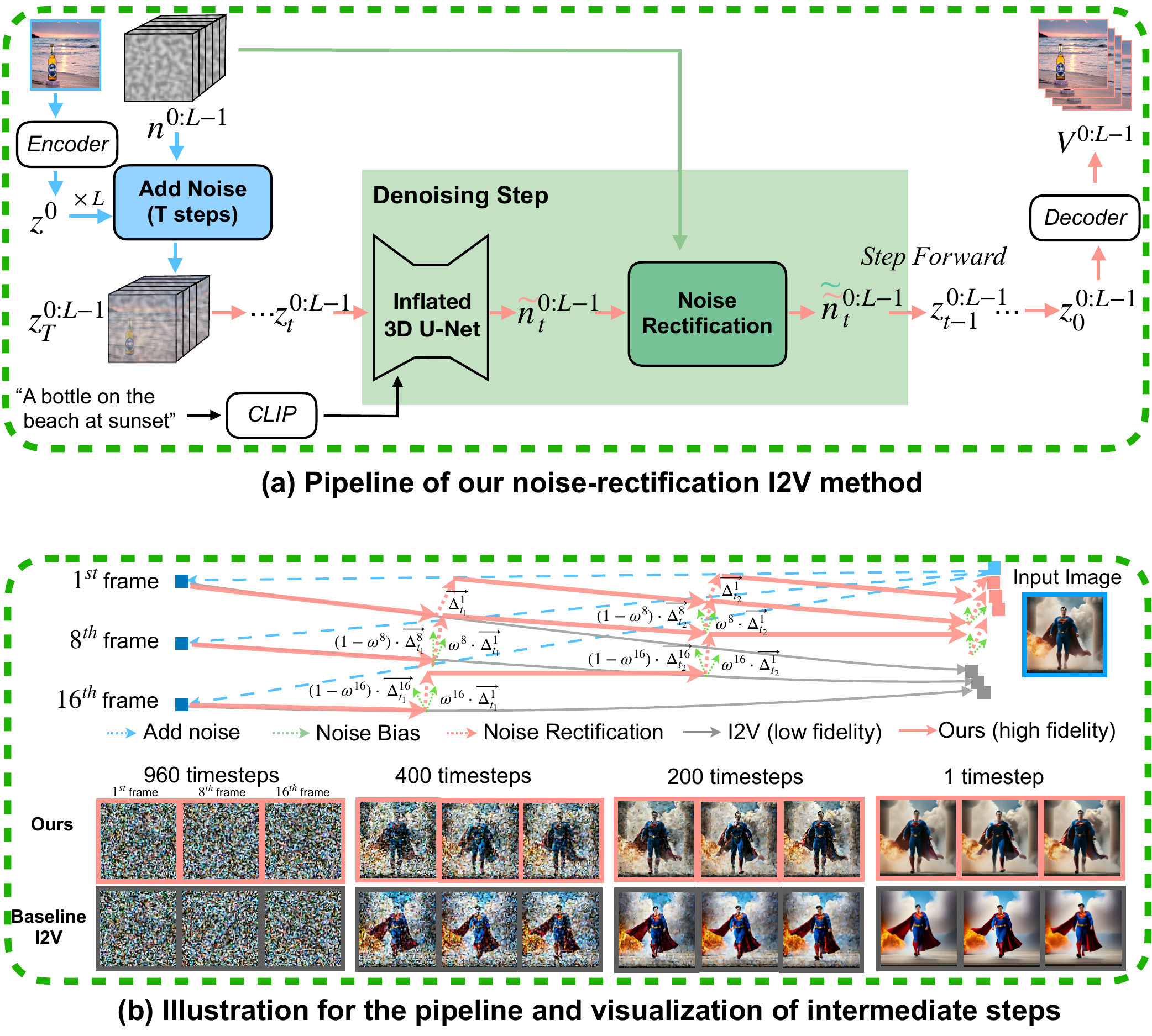}
\caption{The framework of our tuning-free image-to-video method. (a) represents the inference pipeline, where the input image is noised into the initial latent and the predicted noise of the inflated 3D U-Net will be rectified during the denoising process. (b) illustrates the detailed generation process of our method. The visualization of intermediate steps shows our method can effectively refine the denoising direction, making intermediate results closer to the given image.}
\label{fig:framework}
\end{figure*}

Similar to image diffusion generation, video diffusion generation can be regarded as dealing with a batch of images together (see Fig.\ref{fig:image_vs_video}). Recently, video latent diffusion models (VLDM) were also developed upon the text-to-image generation and followed the aforementioned diffusion process, aiming to model the video data from Gaussian noise. Formally, a given video data $V^{0:L-1} \in \mathbb{R}^{L\times 3 \times H \times W}$  will be converted to the latent representation $z_0^{0:L-1} \in \mathbb{R}^{L \times C \times H^{'} \times W^{'}}$ through a VAE encoder~\cite{3_VAE_kingma2013auto}, where $C$ is the number of feature channels. Besides, due to the temporal consistency and content relevance requirements in the video, the VLDM often involves the additional temporal module~\cite{18_VDM_HoSGC0F22,23_ModelScope_wang2023modelscope, 24_Latent-Shift_an2023latentshift, 25_animatediff_guo2023animatediff}, thus inflating the denoising model from 2D U-Net to the 3D U-Net. Through the diffusion process $z_t^{0:L-1} = q(z_0^{0:L-1}, t)$ and reverse process $z_{t-1}^{0:L-1} = p_\theta(z_t^{0:L-1}, t)$, the finally denoised video latent $\bar{z}_0^{0:L-1}$ will be processed via the VAE decoder to generate the video. 

Inspired by the mainstream text-to-video framework, to generate a video from a still image, we also model the video motion with temporal attention in the inflated 3D U-Net \cite{25_animatediff_guo2023animatediff}.  As shown in Fig. \ref{fig:image_vs_video}(b), to improve the computation efficiency, the video’s frame dimension is treated as the batch axis in the forward of the spatial modules, and the video’s spatial dimensions are treated as the batch axis in the forward of the temporal modules.

\section{Method}

\subsection{Enhance Image Condition Analysis}
\label{sec:two_I2V_types}
Although the text-to-video framework can generate a video clip with relatively coherent motion, the semantic content of the generated video is mainly aligned with the given text description at a coarse-grained level. To control the content consistency between the generated video and the reference image, the mainstream I2V works in existing research and the community can be summarized into two basic types (see Fig.\ref{fig:two_I2V_ways}):
One is to incorporate the image condition at the beginning of the reverse process. This approach is mainly inspired by the image generation field like the img2img tasks, such as the image editing task~\cite{15_ILVR_ChoiKJGY21, 16_SDEdit_MengHSSWZE22, 17_Null-text_MokadyHAPC23}, whose idea is to inject the image latent into the initial noise latent. In this way, the reverse denoising process could be implicitly guided towards the direction of the image latent in the latent space. However, this approach can only achieve a resemblance to the given image and there is still a certain gap to high fidelity. A different method involves concatenating the full clean image with the initial noise to introduce more fine-grained details~\cite{34_Seer_gu2023seer, 36_VideoComposer_2023videocomposer, 37_I2VGen-XL_zhang2023i2vgenxl}. While this approach improves fidelity, the entire generation framework must be retrained, leading to low scalability and challenges in integrating with existing pre-trained modules like ControlNet~\cite{12_ControlNet_zhang2023adding}. Another method to enhance image fidelity introduces more image feature signals and conditions into the internal computation of the diffusion model at each timestep~\cite{35_VideoCrafter_chen2023videocrafter1}, such as using various ControlNets~\cite{12_ControlNet_zhang2023adding} and IP-Adapter~\cite{14_IP-Adapter_ye2023ip-adapter}. The image features act as strong supervision to improve the fidelity. However, since feature extraction inevitably loses image details, these approaches tend to learn the overall style or general layout from the original image, making it difficult to achieve high fidelity in terms of fine details.

All the above methods aim to enhance the guidance and control of the initial image in video generation to improve fidelity. However, as shown in Fig.\ref{fig:framework}(b), the denoising process (represented in gray arrow) can not restore the given image even when the initial noisy latents are obtained by adding noise to the given image (represented in dashed blue arrow), we analyzed that the reason why these methods~\cite{35_VideoCrafter_chen2023videocrafter1, 36_VideoComposer_2023videocomposer, 37_I2VGen-XL_zhang2023i2vgenxl,34_Seer_gu2023seer} fail to achieve perfect fidelity lies in the accumulated noise biases during the denoising process, causing the generated frame latents to deviate from the given image latent. In the training process, although the MSE loss function is utilized to make the predicted noise close to the initial input noise, the training process cannot completely achieve a perfect loss of 0. Therefore, there will always be a discrepancy between the predicted noise and the true noise. To further improve fidelity, we draw inspiration from the noise latent and aim to alleviate the noise gap during the denoising process.

\subsection{Noise Rectification Strategy}

Our method includes the ``noising and rectified denoising" process. Similar to~\cite{16_SDEdit_MengHSSWZE22}, our approach starts by injecting the image latent into the initial noise. Without introducing any additional operations, such a setting could generate a coherent video that resembles the given image in the whole style and layout. Taking a different perspective, if the denoising process adopts the known initial noise rather than the predicted biased noise at each timestep, it would result in a video sequence that is entirely faithful but also lacks any motion or dynamics. Therefore, to strike a balance between complete fidelity and dynamics, we propose a noise rectification method. The pipeline of our inference process is shown in Fig.\ref{fig:framework}(a), in some intermediate steps of the denoising process, we rectify the predicted noises by adaptively compensating them with the known initial noise, which is formulated as 

\begin{equation}
    {\color[RGB]{133, 189, 134}\widetilde{\color[RGB]{248, 150, 142}\widetilde{\color{Black}n}}}_t^{0:L-1} = Rectify({\color[RGB]{248, 150, 142}\widetilde{\color{Black}n}}_t^{0:L-1}, n^{0:L-1}, t, \omega^{0:L-1}, \tau),
\end{equation}
where ${\color[RGB]{133, 189, 134}\widetilde{\color[RGB]{248, 150, 142}\widetilde{\color{Black}n}}}_t^{0:L-1}$ denotes the rectified noise at $t^{th}$ timestep, ${\color[RGB]{248, 150, 142}\widetilde{\color{Black}n}}_t^{0:L-1}$ denotes the predicted noise of 3D-UNet, $n^{0:L-1}$ denotes the initial sampled noise that is added to a given image, $\omega^{0:L-1}$ and $\tau$ denote the rectification weight and timestep period. 


\begin{algorithm}[!t]
\renewcommand{\algorithmicrequire}{\textbf{Input:}}
    \renewcommand{\algorithmicensure}{\textbf{Output:}}
    \caption{Noise Rectification for Image-to-Video}
    \begin{algorithmic}[1]
        \REQUIRE{The given image latent $z^{0}$, optional text embedding $c$, video length $L$, rectification weight $\omega ^{0:L-1}$ and timestep period $\tau$.}
        \ENSURE{The generated video latent $z_{0}^{0:L-1}$.}
        \STATE $n^{0:L-1}\sim \mathcal{N}(0,\mathbf{I})$
	\STATE $z_T^{0:L-1} \leftarrow AddNoise(Repeat(z^0), n^{0:L-1}, T)$
	\FOR {$t = T, …, 1$}
          \STATE Predict noise $ {\color[RGB]{248, 150, 142}\widetilde{\color{Black}n}}_t^{0:L-1} = \epsilon_\theta(z_t^{0:L-1}, c, t)$
          \STATE Compute noise gap $\Delta^{0:L-1}_t = {n^{0:L-1} - {\color[RGB]{248, 150,142}\widetilde{\color{Black}n}}_t^{0:L-1}}$
	   \IF {$t$ in $\tau$}
            \STATE Rectify
            $\parbox{\linewidth}{\raggedright${\color[RGB]{133, 189, 134}\widetilde{\color[RGB]{248, 150, 142}\widetilde{\color{Black}n}}}_t^{0:L-1} = {\color[RGB]{248, 150, 142}\widetilde{\color{Black}n}}_t^{0:L-1} + \omega^{0:L-1} \cdot    Repeat(\Delta^0_t) $ \\
            $\hspace{1.1cm}+(1-\omega^{0:L-1}) \cdot \Delta^{0:L-1}_t$
            }$ 
            \ELSE 
            \STATE ${\color[RGB]{133, 189, 134}\widetilde{\color[RGB]{248, 150, 142}\widetilde{\color{Black}n}}}_t^{0:L-1} = {\color[RGB]{248, 150, 142}\widetilde{\color{Black}n}}_t^{0:L-1}$
	   \ENDIF
        \STATE $z_{t-1}^{0:L-1} \leftarrow Sample(z_{t}^{0:L-1}, {\color[RGB]{133, 189, 134}\widetilde{\color[RGB]{248, 150, 142}\widetilde{\color{Black}n}}}_t^{0:L-1})$
	\ENDFOR
        \RETURN $z_{0}^{0:L-1}$
    \end{algorithmic}
\label{algo:noise_recitification}
\end{algorithm}

Concretely, in our noise rectification strategy, the noise ${\color[RGB]{248, 150, 142}\widetilde{\color{Black}n}}_t^{0:L-1}$ predicted by U-Net at each step $t$ is first obtained:
\begin{equation}
    {\color[RGB]{248, 150, 142}\widetilde{\color{Black}n}}_t^{0:L-1} = \epsilon_\theta(z_t^{0:L-1}, c, t),
\end{equation}
where $z_t^{0:L-1}$ is the input latent map at step $t$ and $\epsilon_\theta(\cdot)$ is the denoise model (an inflated 3D U-Net). $c$ and $L$ are the text embedding and video length respectively. Then, we can naturally calculate the noise gap (dubbed $\Delta^{0:L-1}_t$) between the initial sampled noise in our noising process and the noise predicted during the denoising process. 
\begin{equation}
    \Delta^{0:L-1}_t = {n^{0:L-1} - {\color[RGB]{248, 150,142}\widetilde{\color{Black}n}}_t^{0:L-1}}.
\end{equation}

We further calibrate the predicted biased noise, which is the key procedure of our method. By introducing the rectification weight factor $\omega^{0:L-1}$, we balance the first frame noise gap and the subsequent frames' noise gap to obtain the weighted rectification offset, which is then used to frame-wise update the originally predicted noise.
\begin{equation}
\begin{aligned}
   {\color[RGB]{133, 189, 134}\widetilde{\color[RGB]{248, 150, 142}\widetilde{\color{Black}n}}}_t^{0:L-1} &= {\color[RGB]{248, 150, 142}\widetilde{\color{Black}n}}_t^{0:L-1} + \omega^{0:L-1} \cdot Repeat(\Delta^0_t)\\
   &+ (1-\omega^{0:L-1}) \cdot \Delta^{0:L-1}_t,
\end{aligned}
\end{equation}
where $Repeat(\cdot)$ is the broadcasting operation to align the temporal dimension.

\begin{figure*}[t]
\centering
\includegraphics[width=\linewidth,scale=1.0]{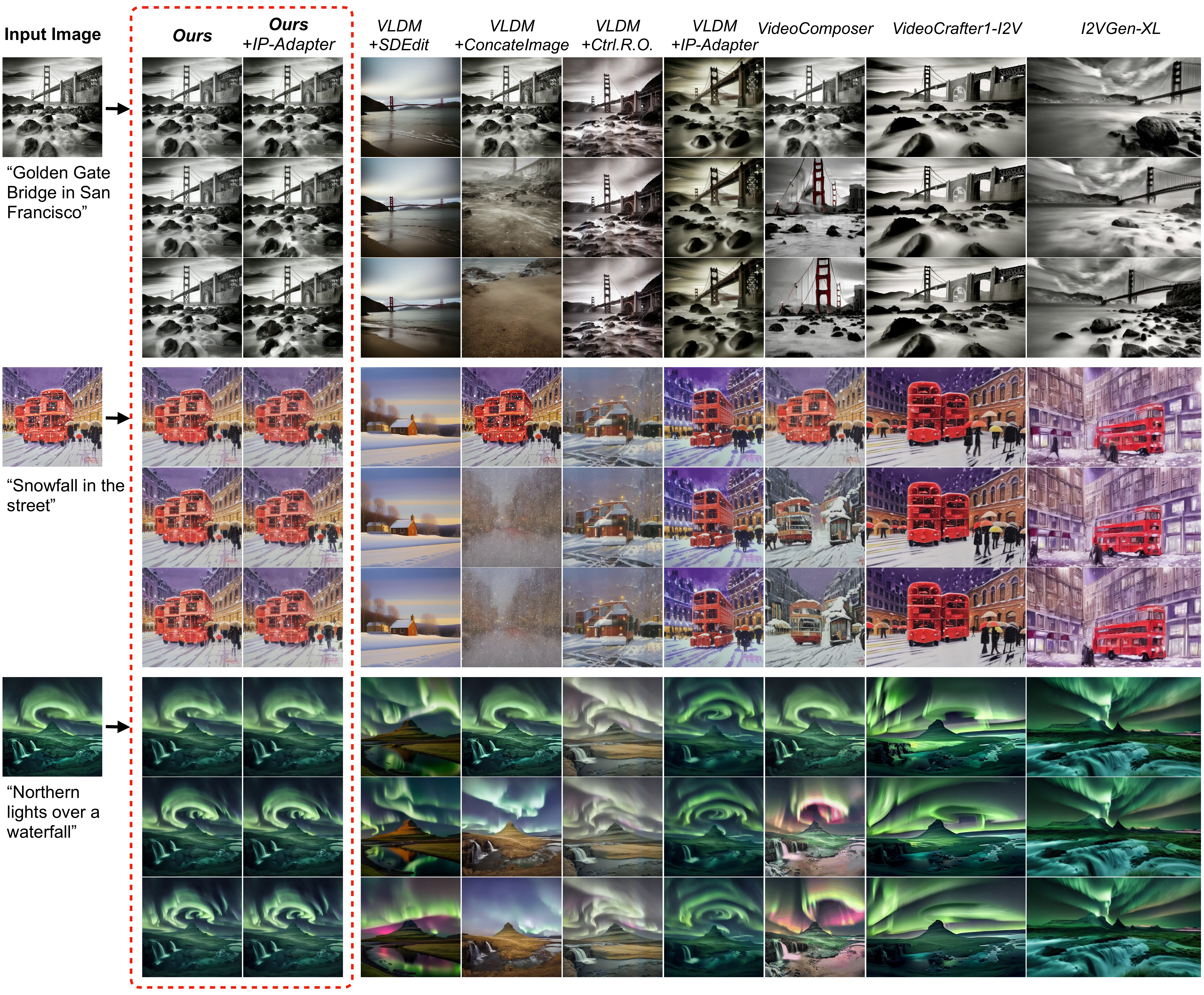}
\caption{Visual comparison with current image-to-video methods. We use AnimateDiff~\cite{25_animatediff_guo2023animatediff} as the VLDM. ``Ctrl.R.O." means ControlNet Reference-Only method~\cite{12_ControlNet_zhang2023adding}. Our method achieves higher fidelity in the video sequences with the given image.}
\label{fig:visual_comparison_result}
\end{figure*}

The whole process of our image-to-video method is detailed in the Algorithm 
\ref{algo:noise_recitification}. Such a noise rectification method is simple but effective. As illustrated in Fig.\ref{fig:framework}(b), through noise rectification (represented in the green arrow), the accumulation noise gap could be effectively alleviated and thus the noisy latent of generated frames are closer to the image latent. In this way, the fine-grained content details of the reference image can be well preserved in the generated video. In addition, to control the retention degree of the reference image, we further introduce a step-adaptive intervention strategy based on noise rectification. Specifically, by adjusting the parameter of rectification steps $\tau$ and weight $\omega^{0:L-1}$, our method could control the fidelity degree of the generated video. It is worth mentioning that our method is tuning-free and can be applied to most current video diffusion models.

\section{Experiments}

\subsection{Experimental Setup}
\textbf{Dataset.} 
We utilized two public datasets WebVid10M~\cite{60_WebVid_bain2021frozen} and LAION-Aesthetic in ~\cite{61_Laion_schuhmann2021laion} to evaluate our method. As for the WebVid10M dataset, we randomly sampled 1000 video-text pairs in proportion to the different categories in its validation subset. For quantitative evaluation, to avoid buffering frames at the beginning of the videos, we selected the 10th frame as the image input for the video generation, along with the video’s text description. As for the LAION-Aesthetic dataset, we also randomly chose 1000 image-text pairs for the qualitative evaluation.

\begin{table*}[!ht]
\centering
\scalebox{1.0}{
\begin{tabular}{lccc}
\toprule[1pt]
Methods & Image fidelity$\text{\ensuremath{\uparrow}}$ & Temporal coherence$\text{\ensuremath{\uparrow}}$ & Video-text alignment$\text{\ensuremath{\uparrow}}$ \\
\midrule[1pt]
VLDM~\cite{25_animatediff_guo2023animatediff} + SDEdit~\cite{16_SDEdit_MengHSSWZE22} & 0.7425 & 0.9888 & \textbf{0.2548} \\
VLDM~\cite{25_animatediff_guo2023animatediff} + ConcateImage & 0.6944 & 0.9427 & 0.2084 \\
VLDM~\cite{25_animatediff_guo2023animatediff} + Ctrl.R.O.~\cite{12_ControlNet_zhang2023adding} & 0.7689 & 0.9919 & 0.2466 \\
VLDM~\cite{25_animatediff_guo2023animatediff} + IP-Adapter~\cite{14_IP-Adapter_ye2023ip-adapter} & 0.7650 & 0.9918 & 0.2287 \\
VideoComposer~\cite{36_VideoComposer_2023videocomposer} & 0.7483  & 0.9352 & 0.2447 \\
VideoCrafter1-I2V~\cite{35_VideoCrafter_chen2023videocrafter1} & 0.7695 & 0.9689 & 0.2206 \\
I2VGen-XL~\cite{37_I2VGen-XL_zhang2023i2vgenxl} & 0.7717 & 0.9560 & 0.2208 \\
\hline 
\textbf{Ours} & 0.7907 & 0.9882 & 0.2517 \\
\textbf{Ours} + IP-Adapter~\cite{14_IP-Adapter_ye2023ip-adapter} & \textbf{0.8042} & \textbf{0.9934} & 0.2405 \\
\bottomrule[1pt] 
\end{tabular}
}
\caption{Quantitative comparison results on the WebVid dataset~\cite{60_WebVid_bain2021frozen}.}
\label{tab:comparison_results}
\end{table*}

\textbf{Evaluation Metrics.}
For the image-to-video generation task, the focus lies on the fidelity and smoothness of the generated videos. Therefore, we assess the generated video from three aspects: image fidelity, temporal coherence, and video-text alignment. Specifically, to evaluate the fidelity between the generated video and the given image, we calculate the CLIP~\cite{8_CLIP_RadfordKHRGASAM21} image similarity for each generated frame and the given image. Considering the temporal consistency in the video, we evaluate the CLIP score between the generated frames. Besides, since the text description is input as a condition, we also calculate the CLIP text-image similarity to evaluate the semantic relevance between the generated video and the text description.

\subsection{Comparisons}
\textbf{Comparison Methods.} We categorized the comparison methods into these two types as Fig.\ref{fig:two_I2V_ways}. One is to incorporate the image condition into the input layer. (1) \textit{SDEdit}~\cite{16_SDEdit_MengHSSWZE22}, a semantic image translation method, which can also be used for I2V tasks by simply adding noises to the given image and then denoising. (2) \textit{ConcateImage}, another simple baseline to concatenate the image condition on initialization noises, which needs to be finetuned to learn the structural information of the given image. Another type of approach is to perform image condition injection at each layer of VLDM. (3) \textit{ControlNet Reference-Only}~\cite{12_ControlNet_zhang2023adding}, an effective way to directly link the attention layer of the VLDM to the reference image. (4) \textit{IP-Adapter}~\cite{14_IP-Adapter_ye2023ip-adapter}, using an additional cross-attention layer for image prompts to achieve semantic and structural preservation. (5) \textit{VideoCrafter1-I2V}~\cite{35_VideoCrafter_chen2023videocrafter1}, similar to IP-Adapter, is another implementation of image prompts injection into the VLDM. Besides, (6) \textit{VideoComposer}~\cite{36_VideoComposer_2023videocomposer} and (7) \textit{I2VGen-XL}~\cite{37_I2VGen-XL_zhang2023i2vgenxl} combine the above two types of ideas for image injection both at the input and middle layers of VLDM.

Benefiting from plug-and-play and tuning-free properties, our method can combine with other image-condition enhancing modules mentioned above. In order to make an intuitive and fair comparison, we conduct our method on both the two above types, denoted as \textit{Ours} and \textit{Ours+IP-Adapter~\cite{14_IP-Adapter_ye2023ip-adapter}}. For fairness, we select AnimateDiff~\cite{25_animatediff_guo2023animatediff} as the pre-trained VLDM.


\begin{figure}[tbp]
\centering
\includegraphics[width=\linewidth,scale=1.0]{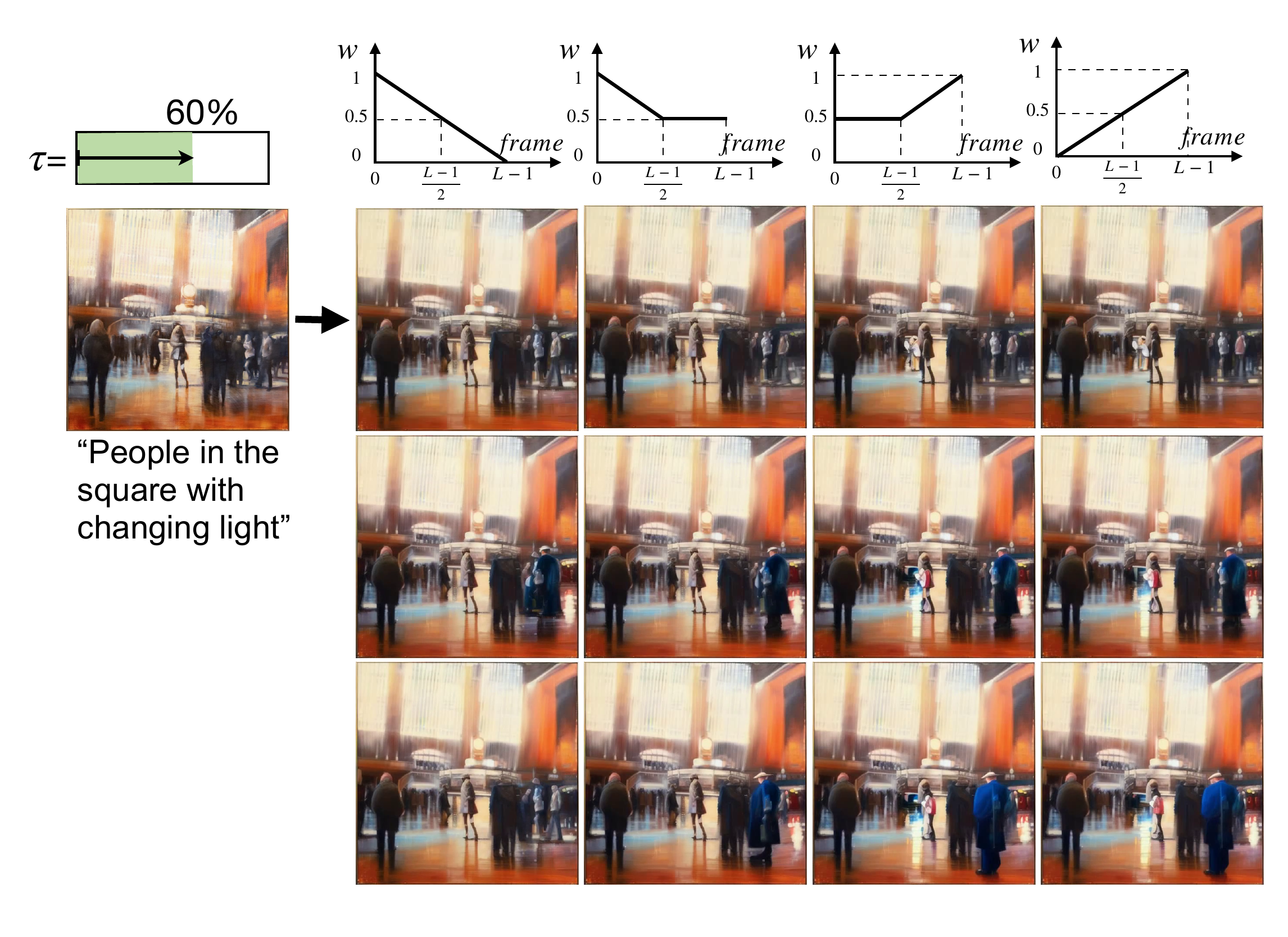}
\caption{Ablation study on the weight of noise rectification. We fix the rectification timestep $\tau$ and change the rectification weights $\omega$ for different frames. }
\label{fig:ablation_on_w}
\end{figure}

\textbf{Qualitative Comparison.}
As shown in Fig.\ref{fig:visual_comparison_result}, the method~\cite{16_SDEdit_MengHSSWZE22} and \textit{ConcateImage} which only incorporate the image condition with the noisy latent at the beginning of the reverse stage are only able to maintain a similar style of the given image. In contrast, those methods~\cite{12_ControlNet_zhang2023adding, 14_IP-Adapter_ye2023ip-adapter,36_VideoComposer_2023videocomposer,35_VideoCrafter_chen2023videocrafter1,37_I2VGen-XL_zhang2023i2vgenxl} that iteratively utilize the image information in the model’s intermediate computation process can preserve more visual features of the given image. In comparison, our method maintains more visual details and achieves high fidelity to the input image. For clearer video samples please refer to the project website.

\begin{figure*}[tbp]
\centering
\includegraphics[width=0.9\linewidth,scale=1.0]{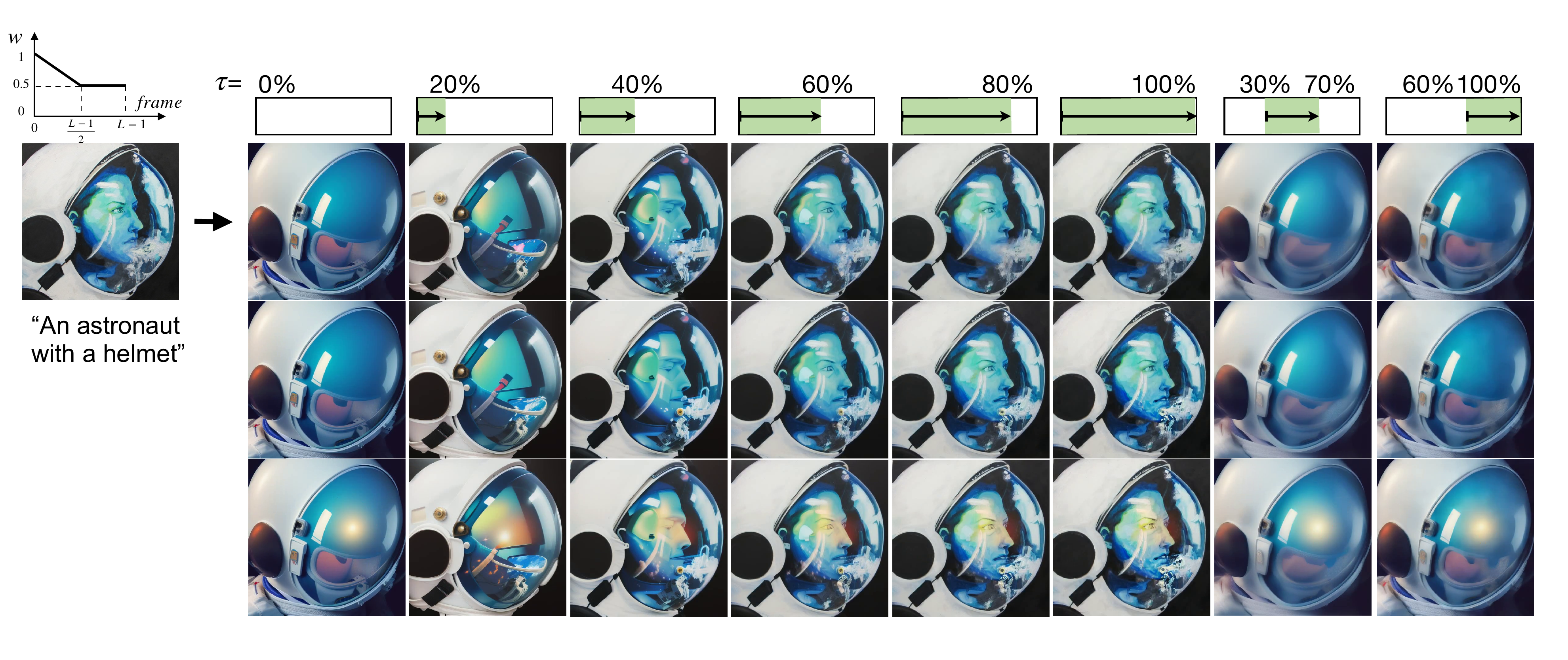}
\caption{Ablation study on the timestep of noise rectification. We fix the rectification weight $\omega$ and the green panels show the the rectification start and end timesteps $\tau$.}
\label{fig:ablation_on_tau}
\end{figure*}

\begin{figure*}[!ht]
\centering
\includegraphics[width=0.9\linewidth,scale=1.0]{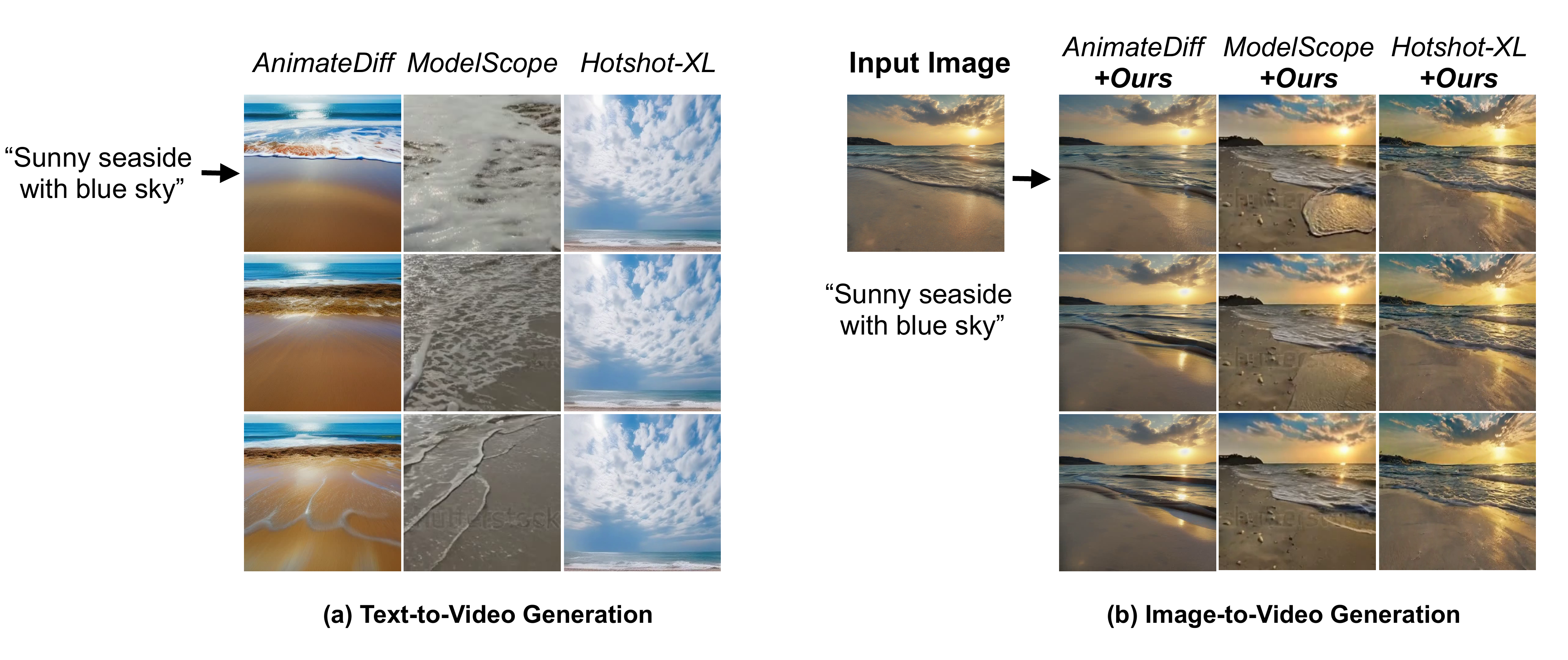}
\caption{A plug-and-play extension of our method on current T2V frameworks to realize I2V. (a) Text-to-video generation results for different T2V models. (b) Different T2V frameworks combined with our method for high-fidelity image-to-video generation.}
\label{fig:extension_to_I2V}
\end{figure*}

\textbf{Quantitative Comparison.}
As shown in Tab.\ref{tab:comparison_results}, our noise rectification method effectively improves the fidelity. Combined with the additional image prompt module~\cite{14_IP-Adapter_ye2023ip-adapter}, our method can obtain the highest video-image fidelity value of 0.8042 and temporal coherence value of 0.9934. Besides, our method still obtains acceptable video-text alignment although we mainly focus on the high fidelity image-to-video task.

\subsection{Ablation Study}
Our method rectifies the predicted noise in the reverse steps and contains two adjustable parameters: rectification weight $\omega$ and rectification timestep $\tau$ as introduced in Algorithm \ref{algo:noise_recitification}. Therefore, we take the ablation study on these two parameters, respectively. Specifically, $\tau=[s, e]$ indicates that noise rectification is performed from $s$ ratio to $e$ ratio of the total timestep.

\textbf{Rectification Weight.} We fix the rectification timestep $\tau=[0\%, 60\%]$ and change the rectification
weights $\omega$ for different frames. The ablation results on $\omega^{i}$ are shown in Fig.\ref{fig:ablation_on_w}, where the plots above the video sequences indicate the rectification weights $\omega^{i}$ for the $i^{th}$ frame. It can be observed that $\omega^{i}$ could affect the fidelity and temporal consistency of subsequent frames. For example, the results in the third or fourth column may result in abrupt changes in the image detail or motion effects. Therefore, we empirically select the setting of the second column for maintaining high fidelity and natural motion.

\textbf{Rectification Timestep.} The rectification timestep period $\tau$ determines in which denoising steps the predicted noise needs to be corrected. As shown in Fig.\ref{fig:ablation_on_tau}, we fix the rectification weights $\omega$ and the green panels show the
rectification start and end timestep. If noise rectification is not performed, i.e. the first column $\tau = [0\%, 0\%]$, the fidelity of the generated video will be poor. Starting from the initial denoising, as the noise rectification period increases (from $\tau=[0\%, 20\%]$ to $\tau=[0\%, 100\%]$), the fidelity will gradually be improved. However, if the rectification only happens on the latter denoising process (i.e.,$\tau=[30\%, 70\%]$ or $\tau=[60\%, 100\%]$), the generated video will still get poor fidelity. These results indicate that accurately predicting noise at the start of the reverse process is crucial for maintaining image fidelity. Considering that a perfect fidelity will scarify the motion intensity, we strike a balance to set $\tau=[0\%, 60\%]$ for all experiments.

\textbf{Extension to More VLDMs.}
Our method utilizes the motion prior of VLDM to model the dynamic motion, which is actually tuning-free and can be adapted to other video diffusion models. To evaluate the extension performance of our method, we selected several recent T2V models and applied our noise rectification method to implement I2V. Besides AnimateDiff~\cite{25_animatediff_guo2023animatediff}, ModelScopeT2V~\cite{23_ModelScope_wang2023modelscope} is a diffusion-based text-to-video model that utilizes the spatio-temporal block to model dynamics. Hotshot-XL~\cite{62_Hotshot-XL_Mullan_Hotshot-XL_2023} is an open-sourced text-to-GIF model developed to work alongside the Stable Diffusion XL (SD-XL) model~\cite{63_SDXL_podell2023sdxl}. We evaluate these three T2V models and extend them to I2V using our plug-and-play noise rectification method. As shown in Fig.\ref{fig:extension_to_I2V}, based on pre-trained motion priors, our method can maintain high fidelity and consistent animation.

\section{Conclusion}
In this work, we propose a simple but effective noise rectification method for image-to-video generation in open domains. We deeply analyze the challenges in I2V and propose a tuning-free approach to ensure high fidelity through a noising and rectified denoising process. Our method is plug-and-play and can be applied to other video latent diffusion models to realize I2V. Experimental results demonstrate the effectiveness of our method. We hope that our method provides a new idea to improve fidelity in the video synthesis field. Notably, our method achieves higher fidelity while losing some motion intensity. Therefore, in future exploration, we will continue to focus on increasing the motion intensity while maintaining high fidelity.
{
    \small
    \bibliographystyle{ieeenat_fullname}
    \bibliography{main}
}


\end{document}


\maketitlesupplementary

\appendix
\section{Comparison Details}

To ensure fairness, we utilized the official open-source code and parameters provided for all comparison methods. 
Specifically, for all models, we employed the CLIP text encoder \cite{8_CLIP_RadfordKHRGASAM21} to get the text embedding, and the VAE model and parameters officially provided by ``stable-diffusion-v1-5" \cite{7_SD_RombachBLEO22} to encode the input image into the latent space. The image $I \in \mathbb{R}^{3\times{H}\times{W}}$ will be converted to the $z^0\in \mathbb{R}^{4\times{\frac{H}{8}}\times{\frac{W}{8}}}$.
(1) \textit{SDEdit}~\cite{16_SDEdit_MengHSSWZE22}, we firstly noised the encoded image latent with the initially sampled noise (using Equation \text{\color{Red}2} in the main text). The noisy latent will be used for the denoising steps in the reverse process.
(2) \textit{ConcateImage}, we concatenated the image latent with the sampled noise through the feature channel, thus resulting the noisy latent $z_T^{0:L-1}\in\mathbb{R}^{L\times8\times{\frac{H}{8}}\times{\frac{W}{8}}}$. Therefore, beyond the 
parameters of VAE and text encoder, the remaining parameters within the framework require fine-tuning. We fine-tuned on WebVid10M Dataset \cite{60_WebVid_bain2021frozen} using the Adam optimizer with a learning rate of 1e-5 for one epoch.
(3) \textit{ControlNet Reference-Only}~\cite{12_ControlNet_zhang2023adding}, we utilized the ``reference-only" processor which directly linked the attention layers of SD to the input image. 
(4) \textit{IP-Adapter}~\cite{14_IP-Adapter_ye2023ip-adapter}, 
we utilized the official ``IP-Adapter Plus" version which extracted the fine-grained image features to get better performance.
(5) \textit{VideoCrafter1-I2V}~\cite{35_VideoCrafter_chen2023videocrafter1}, we directly utilzed the Image2Video model in the released project.
(6) \textit{VideoComposer}~\cite{36_VideoComposer_2023videocomposer}, we both utilized the ``style" and ``single image" mode for the input image.
(7) \textit{I2VGen-XL}~\cite{37_I2VGen-XL_zhang2023i2vgenxl}, 
we used the initial version model which had been open-sourced on the Modelscope platform.
For our methods \textit{Ours} and \textit{Ours+IP-Adapter~\cite{14_IP-Adapter_ye2023ip-adapter}}, as well as $(1)\sim(4)$, we all adopted the pre-trained VLDM (including the ``Realistic Vision V2.0" model and motion module ``mm\_sd\_v15\_v2" in the AnimateDiff~\cite{25_animatediff_guo2023animatediff}. In order to maintain consistency with the official project implementation, the shorter side of input image is resized to 256 for \textit{VideoComposer}~\cite{36_VideoComposer_2023videocomposer} and \textit{I2VGen-XL}~\cite{37_I2VGen-XL_zhang2023i2vgenxl}, 320 for \textit{I2VGen-XL}~\cite{37_I2VGen-XL_zhang2023i2vgenxl} and 512 for other comparison methods. For all generated videos, we all take the length as $L=16$.

\section{Limitations and Society Impacts}
Although our method has effectively improved fidelity in image-to-video generation, there are still some limitations. (1)  Motion intensity. Our method currently is not sufficient for generating larger-scale motion effects. While there is a trade-off between the intensity of motion and fidelity, future work may consider how to increase the magnitude of the motion while preserving the image identity and frame consistency. (2) Video length. Our work, as well as most existing work, is not yet capable of generating longer videos, which poses significant challenges to the maintenance of inter-frame consistency and the ability of storytelling. (3) Controllability. Our method is primarily used for image-to-video generation, aiming to animate still images. However, the current ability to precisely control motion effects in generated videos is still lacking, which may lead to uncertainty in the video output and potential social and ethical risks. In future work, we will focus on enhancing the ability to control the generated video.

{
    \small\bibliographystyle{ieeenat_fullname}
    \bibliography{main}
}
